\documentclass{article}
\usepackage{spconf,amsmath,graphicx}
\usepackage{bbold}
\usepackage[normalem]{ulem}
\usepackage{hyperref}
\usepackage[dvipsnames]{xcolor}
\usepackage{amsmath}
\usepackage{subfigure}
\usepackage{url}
\usepackage{appendix}
\usepackage{enumitem}
\newtheorem{theorem}{Theorem}
\newtheorem{lemma}{Lemma}

\def\x{\textbf{x}}
\def\xhat{\hat{\textbf{x}}}
\def\W{\textbf{W}}
\def\What{\hat{\textbf{W}}}
\def\Wset{\boldsymbol{W}}
\newcommand*{\norm}[2]{\left\|#1\right\|_{#2}}
\newcommand*{\ball}[2]{\mathbb{B}^{\infty}_{#1}(#2)}
\newcommand*{\eps}[1]{\epsilon_{#1}}
\newcommand*{\z}[2]{\textbf{z}^{#1}_{#2}}
\newcommand*{\zhat}[2]{\hat{\textbf{z}}^{#1}_{#2}}

\title{Non-Singular Adversarial Robustness of Neural Networks}
\name{Yu-Lin Tsai$^1$, Chia-Yi Hsu$^1$, Chia-Mu Yu$^1$, Pin-Yu Chen$^2$}
\address{National Chiao Tung University$^1$\\
IBM Research$^2$}

\begin{document}
\ninept
\maketitle
\begin{abstract}
Adversarial robustness has become an emerging challenge for neural network owing to its over-sensitivity to small input perturbations. While being critical, we argue that solving this singular issue alone fails to provide a comprehensive robustness assessment. Even worse, the conclusions drawn from singular robustness may give a false sense of overall model robustness. Specifically, our findings show that adversarially trained models that are robust to input perturbations are still (or even more) vulnerable to weight perturbations when compared to standard models.
In this paper, we formalize the notion of non-singular adversarial robustness for neural networks through the lens of joint perturbations to data inputs as well as model weights. To our best knowledge, this study is the first work considering simultaneous input-weight adversarial perturbations. Based on 
a multi-layer feed-forward neural network model with ReLU activation functions and standard classification loss, we establish error analysis for quantifying the loss sensitivity subject to $\ell_\infty$-norm bounded perturbations on data inputs and model weights. Based on the error analysis, we propose novel regularization functions for robust training and demonstrate improved non-singular robustness against joint input-weight adversarial perturbations.

%\UT{Deep Neural Networks (DNN) has recently achieved great success across many fields, obtaining high accuracy in image classification, speech recognition, translation, etc. However, recent studies shows that it could be at stake when facing adversarial attacks. Data with small perturbations that are imperceptible to human can lead to wrong predictions of machines with high confidence. Many methods were proposed such as adversarial learning to offset this phenomenon so as to raise the so-called 'robustness' of neural networks. We note that most of the papers consider  perturbation in a singular sense ;namely, applied only on weight or input, which is not a comprehensive definition when it comes to robustness of a given neural network.
%In this paper, we start with focus on $\ell_
%\infty$ norm attacks and present the first analysis of error for neural networks with non-negative monotone activation function against simultaneous input and weight perturbations. Secondly, we design brand new loss function based on previous analysis for training a robust neural network under both input and weight perturbations. Lastly, we demonstrate experimental results to validate our findings.}%

\end{abstract}
\begin{keywords}
adversarial example, input perturbation, non-singular adversarial robustness, neural network, weight perturbation
\end{keywords}
\section{Introduction}
Despite recent success achieved by machine learning in a variety of tasks such as object recognition, semantic segmentation, speech recognition and so on, classifiers or predictors remain to perform destructively under the presence of manipulated data subject to perturbations that are imperceptible to human, known as adversarial examples \cite{szegedy2013intriguing,goodfellow2014explaining}. Adversarial examples have been the crux of many attack and defense algorithms tending towards a more adversarially robust model. The notion and mathematical framework of attacks and defences spins off with the development of such algorithms \cite{fawzi2017robustness,biggio2018wild}. 

Specifically,  adversarial examples are often generated from unperturbed data within a norm-ball of radius $\epsilon$. Moreover, the robustness of a model is largely defined as the minimum perturbation that the input could make so as to change a network's correct output \cite{hein2017formal,weng2018evaluating}. In \cite{weng2020towards}, the definition is taken to be modified in order to fit weight (model parameters) perturbation, another type of attack that could cause model to ill-perform. We note that considering input or weight perturbation alone is myopic and incomplete, as it only contributes to \textit{singular} adversarial robustness assessment.
For further reasoning, in Section \ref{sec:experiments} (Fig. \ref{fig:5 models}), we show that models trained under only input perturbation would still suffer when encountering weight perturbation, and vice versa, which suggests that those two singular robustness results poses the risk of offering limited, or even false, sense of the comprehensive model robustness.
%In other words, a truly robust model should be able to withstand perturbations in every possible way. However, papers studying robustness only take either statements into account.

This paper bridges the gap by formalizing \textit{non-singular adversarial robustness} of neural networks and studying simultaneous input-weight perturbations.
%first considering the behaviour of neural networks under simultaneous input and weight perturbations, then offering a margin bound using previous observation.
We develop a novel margin bound analysis on the classification loss for multi-layer neural networks with ReLU activations.
Moreover, based on the analysis, we propose a new loss function towards training robust neural networks against joint input-weight perturbation and validate its effectiveness via empirical experiments. We summarize our contributions as follows.
\begin{itemize}[leftmargin=*]
  \item We study non-singular robustness of neural network using the worst-case bound on pairwise class margin function against joint perturbations in neural networks (Theorems \ref{thm:single_layer} and \ref{thm:multi_layer}). 
  %including two settings, single layer weight-input perturbation and all-layer weight-input perturbation
  \item We propose a theory-driven approach for training non-singular adversarial robust neural networks, including fusing weight perturbation into conventional adversarial training on data inputs \cite{madry2017towards}.
  \item We validate our findings via empirical comparisons with standard and singular adversarial robust neural networks.
\end{itemize}

\section{Related Works}
\label{sec:related}
Recent findings showed that a well-trained neural network can fail catastrophically when adversarial examples are present.
%a well-learned model however accurate is shown in \cite{szegedy2013intriguing} that it can be beaten and fail catastrophically when clean data with imperceptible perturbations are present and was first given the notion of adversarial examples. 
%\cite{goodfellow2014explaining} extended the idea by first considering the range of appropriate perturbations. 
Such adversarial examples can be found by searching within an $\ell_p$ norm-ball of radius $\epsilon$ using gradient-based approaches \cite{goodfellow2014explaining,kurakin2016adversarial_ICLR,moosavi2016deepfool,carlini2017towards,chen2017ead,xu2018structured} or simply using prediction outputs \cite{chen2017zoo,tu2018autozoom,cheng2018query}. 
%and has shown many properties between adversarial examples and neural networks including fast methods of finding such examples. 
Several attack and defense methods were proposed afterwards for studying adversarial robustness. The state-of-the-art robust model presented by \cite{madry2017towards} is composed by a procedure known as adversarial training, where the model weights are updated with the aim of minimizing the worst-case adversarial perturbations, forming a min-max training objective. \cite{wang2019convergence} further proves the convergence of such training process. 
%Despite effective, evaluating the accurate value of the adversarial loss function can be hard and intractable. \cite{raghunathan2018certified}, \cite{wong2018provable} tried to approximate the adversarial loss function under certain relaxation 
%condition and train the model
% upper bound. 

Beyond input perturbations,   \cite{liu2017fault,zhao2019fault} proposed fault-injection attacks which perturb the model parameters stored in memory by physically flipping the logical bits of the memory storage.
\cite{widrow199030,cheney2017robustness} studied weight perturbations applied on the internal architecture for generalization.  \cite{weng2020towards} showed that by taking weight sensitivity into account, the model could maintain its performance after weight quantization. Furthermore, \cite{zhao2020bridging} demonstrated that by taking advantages of mode connectivity of the model's parameters, one could mitigate or preclude the attacks based on weight perturbations. Given the above results, we note that perturbations applied on input or weight has been discussed explicitly but separately, while joint attack remains ambiguous. Meanwhile, it is worth mentioning that adversarial training subject to weight perturbation is not meaningful since the min-max formulation would all be taking place in model's parameter space. In this work, we consider directly when input and weight are both perturbed and prove bounds towards training a non-singular adversarial robust neural network against joint perturbations.

\section{Main Results}
\label{sec:main_results}

In this section we offer an overview of the presentation for our main results as follows.
We first define in Section \ref{sec:notations_and_preliminary} the mathematical notation and preliminary used in this paper. In Section \ref{sec:formalizing_fundamentals}, we introduce the analysis of classification error by first considering a motivating example of a 4-Layer feed-forward neural network, and then further diving into the margin bound of error in the general case. In Section \ref{sec:theory_loss}, we proceed to develop a theory-driven loss function.

\subsection{Notations and Preliminary}
\label{sec:notations_and_preliminary}
We start by offering some mathematical notations used in this paper. Let $[L]$ be the set containing all positive integers smaller than $L$; namely, $[L] := \{1,2,...,L\}$. We write the indicator function as $\mathbb{1}(E)$ which outputs 1 when E occurs and 0 otherwise. As for notations of vectors, we use boldface lowercase letter (e.g. $\x$) and the $i$-th element is marked as $[\x]_{i}$. On the other hand, matrices are denoted by boldface uppercase letter, for example $\W$. Given a matrix $\W \in \mathcal{R}^{m\times n}$, we write its $i$-th row, $j$-th column and $(i,j)$ element as $W_{i,:}$, $W_{;,j}$,and $W_{i,j}$ respectively. The matrix $(\alpha, \beta)$ norm is written as $\norm{\W}{\alpha,\beta}$. In the following sections, we would adopt the notion of vector-induced norm upon mentioning $(\alpha, \beta)$ norm of a given matrix $\W$; namely, we have $ \norm{\W}{\alpha, \beta} = \max_{\x \neq 0}\frac{\norm{\W\x}{\alpha}}{\norm{\x}{\beta}}$. We may use the shorthand notation $\norm{\cdot}{p} := \norm{\cdot}{p,p}$. Furthermore, we use the notion of $\ball{\W}{\epsilon}$ to express an element-wise $\ell_{\infty}$ norm ball for both matrix and vector. Specifically, given a matrix $\W \in \mathcal{R}^{m \times n}$ and vector $\x \in \mathcal{R}^{n}$, we could define the norm ball as $\ball{\W}{\epsilon} := \{\What \hspace{2pt} | \hspace{2pt} |\hat{W}_{i,j} - W_{i,j}| \leq \epsilon, \forall i \in [m], j \in [n]\}$ and $\ball{\x}{\epsilon} := \{\xhat \hspace{2pt} | \hspace{2pt} |[\xhat]_{j} - [\x]_{j}| \leq \epsilon, \forall j \in [n] \} $. 
\newline
\textbf{Preliminary} In order to formally state our results, we start by defining the notion for feed-forward neural networks and laying introduction to a few related quantities. We study multi-class classification problem with number of classes being $K$ in this paper and consider an input vector $\x \in \mathcal{R}^{d}$, an $L$-layer neural network is defined as 
\begin{center}
    $f_{\Wset}(\x) = \W^{L}\rho(\W^{L-1}...\rho(\W^{1}\x)) \in \mathcal{R}^{K}$
\end{center}
with $\Wset$ being the set containing all weight matrices (i.e. $\Wset := \{ \W^{i} \hspace{2pt} | \hspace{2pt} \forall i \in [L] \}$ while $\rho(\cdot)$ stands for non-negative monotone activation functions applied element-wise on a vector and is  assumed to be $1$-Lipschitz which includes popular functions like ReLU and Sigmoid. We further introduce some quantities related to neural networks. The $i$-th component of neural network's output is written as $[f_{\Wset}(\x)]_{i}$ and we denote the pairwise margin, the difference between two classes $i$,$j$ in output of the neural network, as $f^{ij}_{\Wset}(\x):= [f_{\Wset}(\x)]_{i} - [f_{\Wset}(\x)]_{j}$. Finally, we express the output of $k$-th ($k \in [L-1]$) layer given a certain set of matrices $\Wset$ under both unperturbed and input-perturbed setting as $\z{k}{\Wset} := \rho(\W^{k}...\rho(\W^{1}\x)), \W^{m} \in \Wset,  \forall m \in [k] $  and $\zhat{k}{\Wset} :=\rho(\W^{k}...\rho(\W^{1}\xhat))$ where $\xhat \in \ball{\x}{\epsilon_{x}}$ respectively. 

\subsection{Case Study: Joint input and single-layer perturbation}
\label{sec:formalizing_fundamentals}
The sensitivity of neural network in study through the lens of pairwise margin bound, $f^{ij}_{\Wset}(\x)$, especially when $i$ and $j$ corresponds to the top-1 and the second-top prediction of $\x $. We note that the margin bound in the above previous can be utilized as an indicator of robustness given a neural network. For simplicity, we consider a motivating example with 4-Layer neural network and explain the margin bound through the propagation of error under joint perturbation.
We write the neural network $f_{\Wset}(\x)$ as 
\begin{center}
    $f_{\Wset}(\x) = \W^{4}\rho(\W^{3}\rho(\What^{2}\rho(\W^{1}\xhat)))$
\end{center}
where $\W^{i}$ is the weight matrix for the $i$-th layer and assuming that one could perturb any element in the second weight matrix $\W^{2}$ by $\eps{2}$ and any element in the input $\x$ by $\eps{\x}$. Namely, we have $\What^{2} \in \ball{\W^{2}}{\eps{2}}$ and $\xhat \in \ball{\x}{\eps{\x}}$. We define the notion of an error vector $\textbf{e}_{i}$ as the entry-wise error after propagating through the $i$-th layer under only weight perturbation. Consider first the scenario of single-layer weight perturbation, where the second layer is perturbed, then for any input $\x$, since no perturbation happened prior to the second layer, we can take the output after the first layer and derive an upper bound for entries in the error vector $\textbf{e}_{2}$ as 
\begin{align}
    [\textbf{e}_{2}]_{i} := |\hat{W}^{2}_{i,:}\z{1}{\Wset} - W^{2}_{i,:}\z{1}{\Wset}| &\leq \sum_{j} |\hat{W}^{2}_{i,j} - W^{2}_{i,j}||[\z{1}{\Wset}]_{j}| \\ 
    &\leq \sum_{j} \eps{2}|[\z{1}{\Wset}]_{j}| \\
    &= \eps{2} \left\| \z{1}{\Wset} \right\|_{1}
\end{align} 

We next consider each subsequent error vector by the process of propagation. Since no layer after the considered layer is being perturbed, we simply take the magnitude of each element in subsequent weight layer to calculate entries of error vector. Thus, we have that,
\begin{align}
    [\textbf{e}_{3}]_{i} = \sum_{j}|W^{3}_{i,j}||[\textbf{e}_{2}]_{j}| \leq \eps{2}\left\| \z{1}{\Wset} \right\|_{1} \sum_{j}|W^{3}_{i,j}|.
\end{align}
With propagation through layers, we arrive at the final layer and are able to evaluate error induced by perturbations. Recall the pairwise margin bound $f^{ij}_{\Wset}(\x)$, we could derive an upper bound using relative error between entries. Specifically, for any two classes $c_{1}$ and $c_{2}$, we have the relative error in $\textbf{e}_{4}$ as 
\begin{align}
    [\textbf{e}_{4}]_{c_{1}} - [\textbf{e}_{4}]_{c_{2}} &= \sum_{k}{|W^{4}_{c_{1}, k}| - |W^{4}_{c_{2}, k}|)|[\textbf {e}_{3}]_{k}|}\\
    &\leq\eps{2}\left\| \z{1}{\Wset} \right\|_{1} \sum_{k} |W^{4}_{c_{1}, k} - W^{4}_{c_{2}, k}| \sum_{l}|W^{3}_{k,l}| \\
    &\leq \eps{2}\left\| \z{1}{\Wset} \right\|_{1} \max_{k} \norm{W^{3}_{k,:}}{1}\sum_{k} |W^{4}_{c_{1}, k} - W^{4}_{c_{2}, k}| \\
    &= \eps{2}\left\| \z{1}{\Wset} \right\|_{1} \norm{\W^{3}}{\infty} \norm{W^{4}_{c_{1},:} - W^{4}_{c_{2},:}}{1}
\end{align}
From the above example, we could see that the upper bound of relative error would be propagating at the rate of weight matrices' $\ell_\infty$ norm. Thus far, we have derived an upper bound for relative error under single-layer weight perturbation. We proceed to include input perturbation on the basis of single-layer weight perturbation. We denote the error vector $\textbf{e}^{'}_{i}$ as the entry-wise error after propagating through the $i$-th layer under joint perturbation. We can write the first error vector $\textbf{e}^{'}_{1}$ as 
\begin{align}
    [\textbf{e}^{'}_{1}]_{i} := | W^{1}_{i,:}\xhat - W^{1}_{i,:} \x| &\leq \sum_{j} |W^{1}_{i,j}||[\xhat]_{j} - [\x]_{j}| \\
    &\leq \eps{\x} \max_{i} \sum_{j}|W^{1}_{i,j}|\\
    &=  \eps{\x} \norm{\W^{1}}{\infty}
\end{align} The second error vector consists of previous error vector under weight perturbation and the first error vector. Namely, we can write
\begin{align}
    [\textbf{e}_{2}]_{i} &= [\textbf{e}_{2}]_{i} + \sum_{j} |\hat{W}^{2}_{i,j}| |[\textbf{e}_{1}]_{j}| \\ 
    &\leq \eps{2}\norm{\z{1}{\Wset}}{1} + \eps{\x}\norm{\What^{2}}{\infty}\norm{\W^{1}}{\infty}
\end{align}
Since no perturbation is applied afterwards, we could set $\textbf{e}^{'}_{3}$ as 
\begin{align}
    [\textbf{e}^{\prime}_{3}]_{i} &= \sum_{j}|W^{3}_{i,j}||[\textbf{e}^{\prime}_{2}]_{j}| \\ 
    &\leq \underbrace{(\eps{2}\norm{\z{1}{\Wset}}{1} + \eps{\x}\norm{\What^{2}}{\infty}\norm{\W^{1}}{\infty})}_{\text{Denoted as $\eta$}}\sum_{j}|W^{3}_{i,j}|
\end{align}
and proceed to calculate the relative error in $\textbf{e}^{'}_{4}$ between two classes $c_{1}$ and $c_{2}$ as
\begin{align}
   [\textbf{e}^{'}_{4}]_{c_{1}} - [\textbf{e}^{'}_{4}]_{c_{2}} &= \sum_{k}{|W^{4}_{c_{1}, k}| - |W^{4}_{c_{2}, k}|)|[\textbf {e}_{3}^{'}]_{k}|} \\
    &\leq \eta\sum_{k} |W^{4}_{c_{1}, k} - W^{4}_{c_{2}, k}| \sum_{l} |W^{3}_{k,l}| \\
    &\leq \eta\max_{k} \norm{W^{3}_{k,:}}{1} \sum_{k} |W^{4}_{c_{1}, k} - W^{4}_{c_{2}, k}|\\
    &= \eta \norm{\W^{3}}{\infty} \norm{W^{4}_{c_{1},:} - W^{4}_{c_{2},:}}{1}
\end{align}
One can observe that by error propagation, regardless of perturbation type, the error grows with weight matrices $\ell_{\infty}$ norm. Utilizing this concept, we provide two theorem on bounding pairwise margin bound under different perturbation settings. %Due to space limitation, we will provide our full proof in a public web space.
\begin{theorem}[$N$-th layer weight and input joint perturbation]
\label{thm:single_layer}
Let $f_{\Wset}(\x) = \W^{L}(...\rho(\W^{1}\x)...)$ denotes an L-layer neural network and let $f_{\widehat{\Wset}}(\xhat) = \W^{L}(.. \What^{N}...\rho(\W^{1}\xhat)...) $ with $\W^{N} \in \ball{\W^{N}}{\eps{N}}$, and $\xhat \in \ball{\x}{\eps{\x}}$. For any set of perturbed and unperturbed pairwise margin $f^{ij}_{\widehat{\Wset}}(\xhat)$ and $f^{ij}_{\Wset}(\x)$, we have
\begin{align}
f^{ij}_{\widehat{\Wset}}(\xhat) \leq
\begin{cases}
    &\text{if} \hspace{5pt} N \neq L: \nonumber \\  
    &f^{ij}_{\Wset}(\x) + \norm{W^{L}_{i,:} - W^{L}_{j,:}}{1}\Pi_{k=1}^{L-N-1}\norm{\W^{L-k}}{\infty}\times \nonumber \\ 
    &\big\{ \eps{N}\norm{\z{N-1}{\Wset}}{1} + \eps{\x} \Pi_{m = 1}^{N-1} \norm{\W^{m}}{\infty}(\norm{\W^{N}}{\infty}+ d_{N}\eps{N})\big\} \\\\
    &\text{if} \hspace{5pt} N = L: \nonumber \\  
     &f^{ij}_{\Wset}(\x) + \eps{\x}\norm{W^{L}_{i,:} - W^{L}_{j,:}}{1}\Pi_{m=1}^{L-1}\norm{\W^{m}}{\infty} \nonumber \\ 
    &+  2\eps{L}\Pi_{m=1}^{L-1}\norm{\W^{m}}{1}(\norm{\x}{1} + d_{0}\eps{\x})
\end{cases}
\end{align}
where $d_{N}$ stands for the dimensions of $\W^{N}$'s row vector and $d_{0}$ stands for the dimension of input $\x$.
%\\
\end{theorem}
\textit{Proof}:
See Appendix \ref{appx_a.1}.

%\begin{lemma}[final layer weight and input perturbation]
%Consider the case when N = L in the above Theorem \ref{thm:single_layer}, we then have for any set of perturbed and unperturbed pairwise margin,
%\begin{align}
%    &f^{ij}_{\widehat{\Wset}}(\xhat) \leq  f^{ij}_{\Wset}(\x) + \eps{\x}\norm{W^{L}_{i,:} - W^{L}_{j,:}}{1}\Pi_{m=1}^{L-1}\norm{\W^{m}}{\infty} \nonumber \\ 
%    &+  2\eps{L}\Pi_{m=1}^{L-1}\norm{\W^{m}}{1}(\norm{\x}{1} + d_{0}\eps{\x})
%\end{align}
%where $d_{0}$ stands for the dimension of input $\x$

%\end{lemma}
%\textsl{Proof}: Proof will be provided as a supplementary material

\begin{theorem}[all-layer and input joint perturbation]
\label{thm:multi_layer}
Let $f_{\Wset}(\x)= \W^{L}\rho(..\rho(\W^{1}\x)...)$ denotes an L-layer neural network and let $f_{\widehat{\Wset}}(\xhat) = \What^{L}\rho(..\rho(\What^{1}\xhat)...)$ with $\What^{m} \in \ball{\W^{m}}{\eps{m}}, \forall m \in [L]$ and $\xhat \in \ball{\x}{\eps{x}}$, furthermore, let $\xi$ be the set containing possible perturbations, i.e. $\xi := \eps{\x}\cup \{\eps{m}\}_{m=1}^{L} $ and $d_{m}$ representing the dimension of matrix $\W^{m}$'s row vector, then for any set of pairwise margin bound between natural and joint perturbed settings, we have
\begin{align}
    f^{ij}_{\widehat{\Wset}}(\xhat) \leq f^{ij}_{\Wset}(\x) + \tau^{ij}_{\Wset}(\xi) + \zeta^{ij}_{\Wset}(\x , \xi)
\end{align}
where $\tau^{ij}_{\Wset}(\xi)$ can be expressed as
\begin{align}
    \tau^{ij}_{\Wset}(\xi) = \eps{\x}\bigg(\norm{W^{L}_{i,:} - W^{L}_{j,:}}{1} + 2d_{L}\eps{L}\bigg)\Pi_{m = 1}^{L-1}(\norm{\W^{m}}{\infty} + d_{m}\eps{m})
\end{align}
while $\zeta^{ij}_{\Wset}(\x, \xi)$ possesses the following form
\begin{align}
    &\zeta^{ij}_{\Wset}(\x, \xi) := \left\|W^{L}_{i,:} - W^{L}_{j,:} \right\|_{1} \bigg\{ \epsilon_{1}\left\| \textbf{x} \right\|_{1} \Pi_{l = 1}^{L-2} \left\|(\textbf{W}^{L-l})\right \|_{\infty} \nonumber \\
    &+  \sum_{k = 1}^{L-3} \big (\Pi_{m = k+2}^{L-1} \left\| \textbf{W}^{m} \right \|_{\infty} \big)\epsilon_{k+1} \left\| \textbf{h}^{k^*} \right\|_{1}
    + \epsilon_{L-1} \left\| \textbf{h}^{{L-2}^*} \right\|_{1} \bigg\} \nonumber \\ 
    &+ 2\epsilon_{L}\left\| \textbf{h}^{{L-1}^*} \right\|_{1} \nonumber \\ 
    &\text{where} \hspace{5pt} \textbf{h}^{k^*} = \rho(\textbf{W}^{k^*}...\rho(\textbf{W}^{1^*}
    \textbf{x}) \nonumber \\  
    &\text{with}
    \begin{cases}
        W^{m^*}_{i,j} = W^{m}_{i,j} + \epsilon_{m},\ \forall i, j \ \text{and}~\forall m \in [L] \setminus \{1\} \vspace{5pt} \\
        W^{1^*}_{i,j} = W^{1}_{i,j} + sgn([\textbf{x}]_{j})\hspace{2pt}\epsilon_1, \ \forall i, j
    \end{cases}
\end{align}
\end{theorem}
\textit{Proof}:
See Appendix \ref{appx:a.2}.

\subsection{Theory-inspired loss towards non-singular robustness}
\label{sec:theory_loss}
With our theoretical insights on margin bound, we now propose new regularization function towards training a non-singular adversarial robust neural network. Specifically, consider the new loss function in the following form:
\begin{align}
\label{eq_new_loss}
    \ell^{\prime}(f_{\Wset}(\x), y) &= \ell_{\text{cls}}(f_{\Wset}(\x), y) + \alpha \max_{y^{\prime} \neq y}\{ \tau^{y^{\prime}y}_{\Wset}(\xi) \} \nonumber \\
    &~~~+ \beta \max_{y^{\prime} \neq y} \{ \zeta^{y^{\prime} y}_{\Wset}(\x, \xi)\}
\end{align}
where the first term $\ell_{\text{cls}}$ corresponds to standard (or adversarial)  classification loss, while
the second and third term regularizes perturbation sensitivity to input and weight space with nonnegative coefficients $\alpha$ and $\beta$, respectively. They are inspired from Theorem \ref{thm:multi_layer} and can be interpreted as the maximum error on pairwise margin induced by joint input-weight perturbations. Specifically, each regularizer alone corresponds to singular sensitivity, while their mixture governs non-singular adversarial robustness.
%Traditional classification loss functions aims to increase model accuracy by widening pairwise margin; however, certain joint perturbation may still incur huge error and thus render the model useless. To address this issue, we propose to train under mixed regularization objective which balances between standard performance and non-singular robustness. 
%In the above equation, the first term comes form standard classification task while we interpret the second and third term as the maximum error on pairwise margin induced by joint perturbations on weights and input. Specifically, the third term is a result of maximum error considered only the weight perturbation scenario while the second term focuses on error caused by input perturbation on the basis of former scenario.

\begin{figure*}[t]
    \centering
    \subfigure[Standard Model (AUC=12.18)]{
    \includegraphics[width=0.32\textwidth]{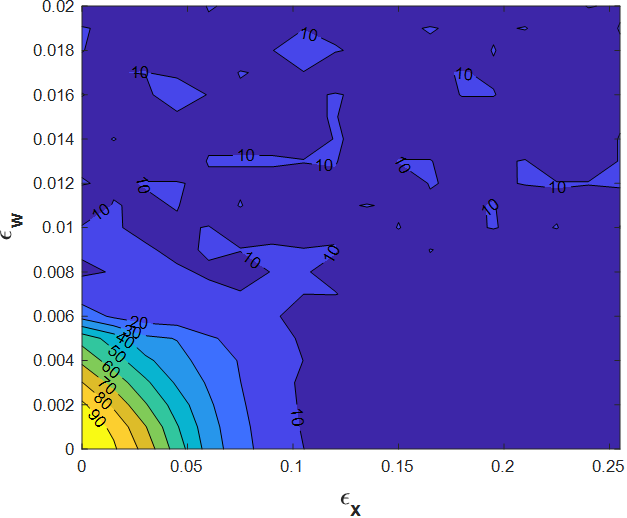}}
    \subfigure[Weight Perturb (AUC=15.07)]{
    \includegraphics[width=0.32\textwidth]{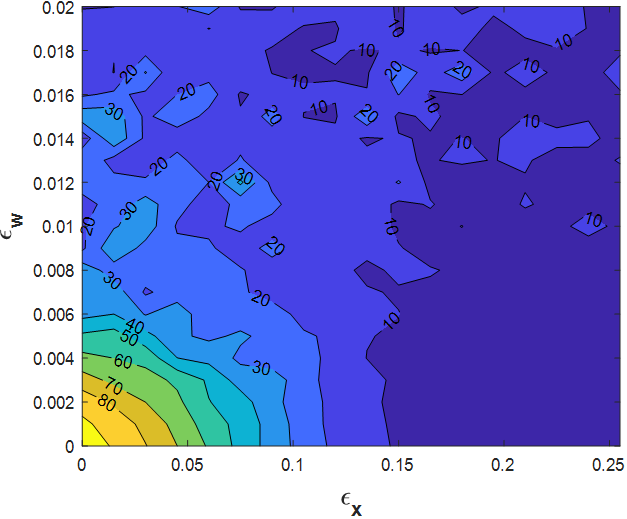}}
    \subfigure[AT (AUC=20.37)]{
    \includegraphics[width=0.32\textwidth]{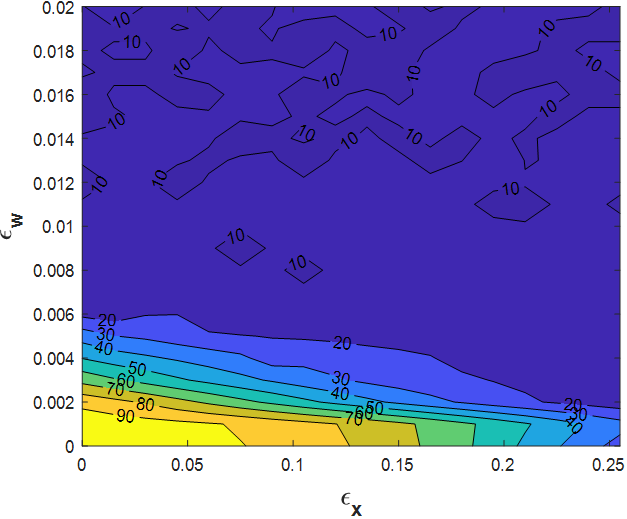}}
    \subfigure[AT+$\beta$-1 (AUC=21.51)]{
    \includegraphics[width=0.32\textwidth]{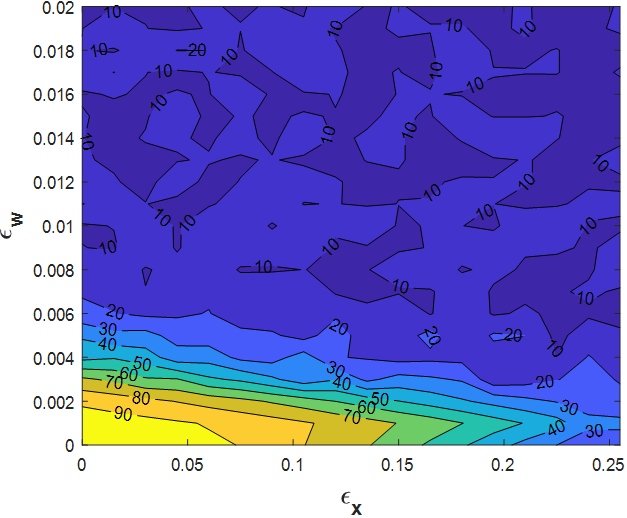}}
    \subfigure[AT+$\beta$-2 (AUC=25.23)]{
   \includegraphics[width=0.32\textwidth]{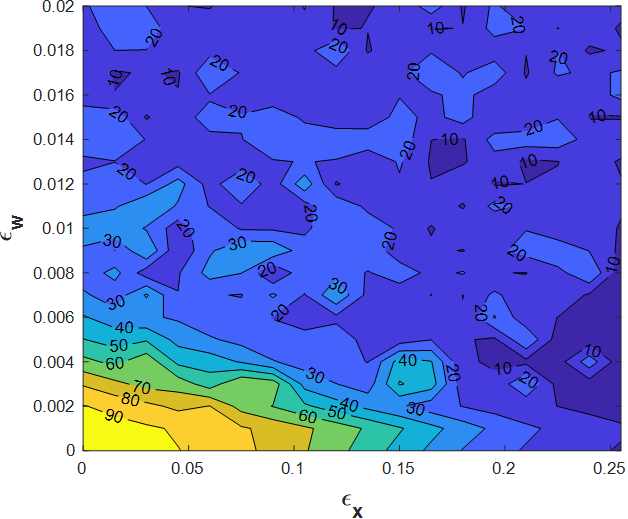}}
    \subfigure[JIWP (AUC=17.74)]{
    \includegraphics[width=0.32\textwidth]{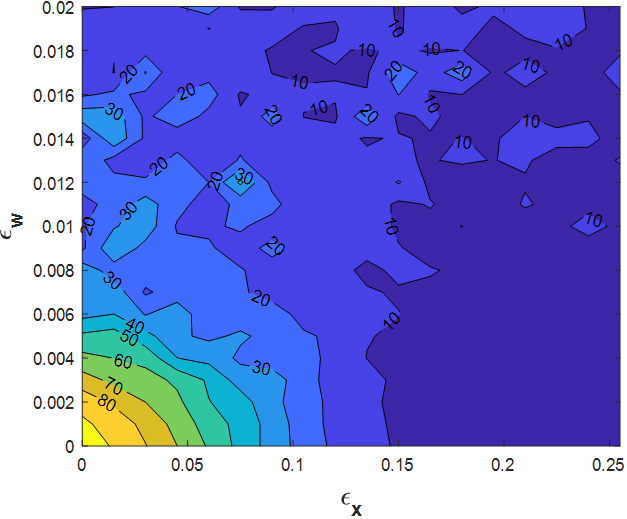}}
    \vspace{-4mm}
    \caption{Comparison of test accuracy contour of neural networks under joint input-weight PGD attack (100 steps) with varying input ($\epsilon_{x}$) and weight ($\epsilon_{w}$) perturbation levels. AUC refers to the area under curve scores. Comparing to the  the standard model (a), singular robust models (b) and (c) have comparable or even worse robustness under their respective untrained perturbation type. Non-singular robust models using our proposed regularization function, including (d), (e) and (f), show significantly better AUC scores.
    %The input/weight robustness was low when we took singular perturbation or none into consideration shown in (a), (b) and (c). Taking account of both input and weight perturbation was shown in (d), (e) and (f) against input-weight PGD attack have brilliant results .
    }
    \label{fig:5 models}
        \vspace{-2mm}
\end{figure*}

\section{Experiments}
\label{sec:experiments}

\subsection{Experiment Setup}
We used the MNIST image classification dataset containing 10 hand-written digit categories. We trained neural network models with four dense layers (number of neurons are 128-64-32-10) and the ReLU activation function without the bias term. For comparison, five different training methods using the training loss in (\ref{eq_new_loss}) are presented in our experiments: (i) Standard Model, (ii) Weight Perturb, (iii) Adversarial Training (AT) \cite{madry2017towards}, (iv) Adversarial Training with additional $\beta$-term regularizaiton  (AT+$\beta$), and (v) Joint Input-Weight Perturb (JIWP). 
%To get better results, we have already tuned the weight perturbation level $\epsilon_{w}$ and input perturbation level $\epsilon_{x}$. 
%Instead of discussion values of $\alpha$ and $\beta$, we set they are the same and tuned them for all models. 
To obtain reasonable accuracy on the unperturbed testing data, we have tuned the models with weight and input perturbation levels $\epsilon^{\text{train}}_{w}$ and  $\epsilon^{\text{train}}_{x}$ and regularization coefficients $\alpha$ and $\beta$ for each model.
For the standard model, we used the cross entropy (CE) for $\ell_{\text{cls}}$ with $\alpha=\beta=0$. For the weight perturb model,
we used the CE loss function $\ell^{\prime}$ ($\alpha=\beta=0.25$) with $\epsilon^{\text{train}}_{w}=0.01$ and $\epsilon^{\text{train}}_{x}=0$. For AT, we followed the same min-max training setting with CE loss as in \cite{madry2017towards} and set $\epsilon^{\text{train}}_{x} =0.09$, and $\alpha=\beta=0$. We trained two AT+$\beta$ models with $\alpha=0$ and using $(\beta,\epsilon^{\text{train}}_{x})= (0.0003,0.08)$ (AT+$\beta$-1) and $(\beta,\epsilon^{\text{train}}_{x})= (0.005,0.03)$ (AT+$\beta$-2), respectively. For the JIWP model, we set $\alpha = \beta = 0.02$, $\epsilon^{\text{train}}_{w}=0.02$ and $\epsilon^{\text{train}}_{x}=0.3$. Except for AT, AT+$\beta$-1 and AT+$\beta$-2, we used Adam optimizer with initial learning rate $10^{-4}$, a batch size of 50, and 300 training epochs. 

\subsection{Performance Evaluation}
For non-singular robustness evaluation, we generalize  the projected gradient descent (PGD) attack \cite{madry2017towards} for input perturbation to joint input-weight perturbation, by simultaneously computing the signed gradient of the CE loss with respect to the data input and the model weight, clipping the perturbation within their respective $\ell_\infty$ ball constraints, and iterate this process for 100 steps with step sizes $\alpha_{\boldsymbol{X}}=0.01$ and $\alpha_{\boldsymbol{W}}=0.0005$.
%In order to evaluate robustness against adversarial joint input-weight perturbation, we designed for joint input-weight perturbation with adjusting the projected gradient descent (PGD) attack originally which is called input-weight PGD attack.  
We describe this joint PGD attack as follows.
Given an input $\boldsymbol{X}$ and a trained neural network weight $\boldsymbol{W}$, the perturbed weight $\widetilde{\boldsymbol{W}}$ and input $\widetilde{\boldsymbol{X}}$ are crafted by iterative gradient ascent using the sign of gradient of the CE loss marked as $ \text{sgn} ( \nabla_{\boldsymbol{W,X}} \ell_{cls}(f_{\widetilde{\boldsymbol{W}}}(\widetilde{\boldsymbol{X}}),y) )$. 
The attack iteration with step sizes $\alpha_{\boldsymbol{W}}$ of weight and $\alpha_{\boldsymbol{X}}$ of input is formalized as
{ \footnotesize
\begin{align*}
&\widetilde{\boldsymbol{W}}^{(0)} = \boldsymbol{W},\\ &\widetilde{\boldsymbol{W}}^{(t+1)} = \text{Clip}_{\boldsymbol{W}, \epsilon_{w}} \left\{ \widetilde{\boldsymbol{W}}^{(t)} + \alpha_{\boldsymbol{W}}  \text{sgn} (\nabla_{\boldsymbol{W,X}}\ell_{cls}(f_{\widetilde{\boldsymbol{W}}^{(t)}}(\widetilde{\boldsymbol{X}}^{(t)} ),y) ) \right\} 
\end{align*} }
%\vspace{-6mm}
{ \footnotesize
\begin{align*}
&\widetilde{\boldsymbol{X}}^{(0)} = \boldsymbol{X},\\ &\widetilde{\boldsymbol{X}}^{(t+1)} = \text{Clip}_{\boldsymbol{X}, \epsilon_{x}} \left\{ \widetilde{\boldsymbol{X}}^{(t)} + \alpha_{\boldsymbol{X}} \; \text{sgn} (\nabla_{\boldsymbol{W,X}}\ell_{cls}(f_{\widetilde{\boldsymbol{W}}^{(t)}}(\widetilde{\boldsymbol{X}}^{(t)}),y) ) \right\} 
\end{align*} 
}
% \begin{figure}[t]
%     \centering
%     \includegraphics[width=0.3\textwidth]{images/fix_input.png}
%     \vspace{-2mm}
%     \caption{Comparison of area under curve (AUC) across $\epsilon_w$ ranging from 0 to 0.02 at different $\epsilon_x$ values. }
%     \label{fig:fixed input}
% \end{figure}

Fig.\ref{fig:5 models}
demonstrates the non-singular robustness performance for each model. The standard model (a) is vulnerable to both weight and input perturbations. Singular robust models (b) and (c) are only robust to the seen perturbation type, while they only have comparable or even worse robustness against unseen  perturbation type. For example, AT (model (c)) is only trained on input perturbation and is observed to be less robust under weight perturbation compared to the standard model (a). Similarly, the robustness of weight perturb model (b) to input perturbation is only slightly better than the standard model. The results suggest the insufficiency of singular robustness analysis.
Comparing the area under curve (AUC) score of test accuracy, non-singular robust models (bottom row, (d)-(f)) using our proposed loss significantly  outperform standard and singular robust models (top row). The AUC of best AT+$\beta$ model (e) improves that of AT by about 24\%, validating the effectiveness of our proposed regularizer. AT+$\beta$ also attains better AUC than JIWP, suggesting that min-max training is crucial to non-singular robustness.
%The weight perturb model can improve the robustness of weight perturbation more than input perturbation. Similarly, adversarial training has a stronger resistance of input perturbation, but the resistance of weight perturbation even weaker than the standard model. It is not focus on certain perturbation that can maintain high accuracies under larger $\epsilon_x$ and $\epsilon_w$ 
%simultaneously shown in Fig.\ref{fig:5 models} (d) and (e). Moreover, 
%Fig. \ref{fig:fixed input} shows that is a fundamental robustness trade-off between input and weight perturbation, especially for large perturbation levels. Specifically, by inspecting the AUC (over weight perturbation levels) at a fixed input perturbation level, the best non-singular adversarial robust model may vary, especially for large input perturbation levels. 
%It is proper non-zero values of $\epsilon_{x}$ and $\epsilon_{w}$ that we could get the highest AUC. On the other hand, we only concern about weight or input perturbations bringing about reductions in performance.

\section{Conclusion}
\label{sec:conclusion}
In this paper, we analyze the robustness of pairwise class margin for neural networks against joint input-weight perturbations. A theory-inspired regularizer is proposed towards training comprehensive robust neural networks. 
Empirical results against joint input-weight perturbations show that singular robust models can give a false sense of overall robustness, while our proposal can significantly improve non-singular adversarial robustness and offer thorough evaluation.

\bibliographystyle{IEEEbib}
\bibliography{ref}

\begin{thebibliography}{10}

\bibitem{szegedy2013intriguing}
Christian Szegedy, Wojciech Zaremba, Ilya Sutskever, Joan Bruna, Dumitru Erhan,
  Ian Goodfellow, and Rob Fergus,
\newblock ``Intriguing properties of neural networks,''
\newblock {\em International Conference on Learning Representations}, 2014.

\bibitem{goodfellow2014explaining}
Ian~J Goodfellow, Jonathon Shlens, and Christian Szegedy,
\newblock ``Explaining and harnessing adversarial examples,''
\newblock {\em arXiv preprint arXiv:1412.6572}, 2014.

\bibitem{fawzi2017robustness}
Alhussein Fawzi, Seyed-Mohsen Moosavi-Dezfooli, and Pascal Frossard,
\newblock ``The robustness of deep networks: A geometrical perspective,''
\newblock {\em IEEE Signal Processing Magazine}, vol. 34, no. 6, pp. 50--62,
  2017.

\bibitem{biggio2018wild}
Battista Biggio and Fabio Roli,
\newblock ``Wild patterns: Ten years after the rise of adversarial machine
  learning,''
\newblock {\em Pattern Recognition}, vol. 84, pp. 317--331, 2018.

\bibitem{hein2017formal}
Matthias Hein and Maksym Andriushchenko,
\newblock ``Formal guarantees on the robustness of a classifier against
  adversarial manipulation,''
\newblock in {\em Advances in Neural Information Processing Systems}, 2017, pp.
  2263--2273.

\bibitem{weng2018evaluating}
Tsui-Wei Weng, Huan Zhang, Pin-Yu Chen, Jinfeng Yi, Dong Su, Yupeng Gao,
  Cho-Jui Hsieh, and Luca Daniel,
\newblock ``Evaluating the robustness of neural networks: An extreme value
  theory approach,''
\newblock {\em International Conference on Learning Representations}, 2018.

\bibitem{weng2020towards}
Tsui-Wei Weng, Pu~Zhao, Sijia Liu, Pin-Yu Chen, Xue Lin, and Luca Daniel,
\newblock ``Towards certificated model robustness against weight
  perturbations.,''
\newblock in {\em Proceedings of the AAAI Conference on Artificial
  Intelligence}, 2020, pp. 6356--6363.

\bibitem{madry2017towards}
Aleksander Madry, Aleksandar Makelov, Ludwig Schmidt, Dimitris Tsipras, and
  Adrian Vladu,
\newblock ``Towards deep learning models resistant to adversarial attacks,''
\newblock {\em International Conference on Learning Representations}, 2018.

\bibitem{kurakin2016adversarial_ICLR}
Alexey Kurakin, Ian Goodfellow, and Samy Bengio,
\newblock ``Adversarial machine learning at scale,''
\newblock {\em International Conference on Learning Representations}, 2017.

\bibitem{moosavi2016deepfool}
Seyed-Mohsen Moosavi-Dezfooli, Alhussein Fawzi, and Pascal Frossard,
\newblock ``Deepfool: a simple and accurate method to fool deep neural
  networks,''
\newblock in {\em IEEE Conference on Computer Vision and Pattern Recognition},
  2016, pp. 2574--2582.

\bibitem{carlini2017towards}
Nicholas Carlini and David Wagner,
\newblock ``Towards evaluating the robustness of neural networks,''
\newblock in {\em IEEE Symposium on Security and Privacy}, 2017, pp. 39--57.

\bibitem{chen2017ead}
Pin-Yu Chen, Yash Sharma, Huan Zhang, Jinfeng Yi, and Cho-Jui Hsieh,
\newblock ``{EAD}: elastic-net attacks to deep neural networks via adversarial
  examples,''
\newblock in {\em Proceedings of the AAAI Conference on Artificial
  Intelligence}, 2018, pp. 10--17.

\bibitem{xu2018structured}
Kaidi Xu, Sijia Liu, Pu~Zhao, Pin-Yu Chen, Huan Zhang, Quanfu Fan, Deniz
  Erdogmus, Yanzhi Wang, and Xue Lin,
\newblock ``Structured adversarial attack: Towards general implementation and
  better interpretability,''
\newblock {\em International Conference on Learning Representations}, 2019.

\bibitem{chen2017zoo}
Pin-Yu Chen, Huan Zhang, Yash Sharma, Jinfeng Yi, and Cho-Jui Hsieh,
\newblock ``{ZOO}: Zeroth order optimization based black-box attacks to deep
  neural networks without training substitute models,''
\newblock in {\em ACM Workshop on Artificial Intelligence and Security}, 2017,
  pp. 15--26.

\bibitem{tu2018autozoom}
Chun-Chen Tu, Paishun Ting, Pin-Yu Chen, Sijia Liu, Huan Zhang, Jinfeng Yi,
  Cho-Jui Hsieh, and Shin-Ming Cheng,
\newblock ``Autozoom: Autoencoder-based zeroth order optimization method for
  attacking black-box neural networks,''
\newblock in {\em Proceedings of the AAAI Conference on Artificial
  Intelligence}, 2019, vol.~33, pp. 742--749.

\bibitem{cheng2018query}
Minhao Cheng, Thong Le, Pin-Yu Chen, Jinfeng Yi, Huan Zhang, and Cho-Jui Hsieh,
\newblock ``Query-efficient hard-label black-box attack: An optimization-based
  approach,''
\newblock {\em International Conference on Learning Representations}, 2019.

\bibitem{wang2019convergence}
Yisen Wang, Xingjun Ma, James Bailey, Jinfeng Yi, Bowen Zhou, and Quanquan Gu,
\newblock ``On the convergence and robustness of adversarial training.,''
\newblock in {\em ICML}, 2019, vol.~1, p.~2.

\bibitem{liu2017fault}
Yannan Liu, Lingxiao Wei, Bo~Luo, and Qiang Xu,
\newblock ``Fault injection attack on deep neural network,''
\newblock in {\em 2017 IEEE/ACM International Conference on Computer-Aided
  Design (ICCAD)}. IEEE, 2017, pp. 131--138.

\bibitem{zhao2019fault}
Pu~Zhao, Siyue Wang, Cheng Gongye, Yanzhi Wang, Yunsi Fei, and Xue Lin,
\newblock ``Fault sneaking attack: A stealthy framework for misleading deep
  neural networks,''
\newblock in {\em 2019 56th ACM/IEEE Design Automation Conference (DAC)}. IEEE,
  2019, pp. 1--6.

\bibitem{widrow199030}
Bernard Widrow and Michael~A Lehr,
\newblock ``30 years of adaptive neural networks: perceptron, madaline, and
  backpropagation,''
\newblock {\em Proceedings of the IEEE}, vol. 78, no. 9, pp. 1415--1442, 1990.

\bibitem{cheney2017robustness}
Nicholas Cheney, Martin Schrimpf, and Gabriel Kreiman,
\newblock ``On the robustness of convolutional neural networks to internal
  architecture and weight perturbations,''
\newblock {\em arXiv preprint arXiv:1703.08245}, 2017.

\bibitem{zhao2020bridging}
Pu~Zhao, Pin-Yu Chen, Payel Das, Karthikeyan~Natesan Ramamurthy, and Xue Lin,
\newblock ``Bridging mode connectivity in loss landscapes and adversarial
  robustness,''
\newblock in {\em International Conference on Learning Representations}, 2020.

\end{thebibliography}

\appendix
\onecolumn
\section{Proof of Theorems}
\subsection{Theorem 1: Single-Layer Bound}\label{appx_a.1}
We shall first prove when $N \neq L$ and follow similar reasoning to prove the case when $N = L$. Consider the difference between set of pairwise margin $f^{ij}_{\What}(\x) - f^{ij}_{\W}(\x)$, we have
\begin{align}
    &f^{ij}_{\widehat{\boldsymbol{W}}}(\xhat) - f^{ij}_{\boldsymbol{W}}(\textbf{x}) \nonumber \\
    &= f^{ij}_{\hat{\Wset}}(\xhat) - f^{ij}_{\hat{\Wset}}(\x) + f^{ij}_{
\hat{\Wset}}(\x) - f^{ij}_{\Wset}(\x) \\
    &\overset{(a)}{\leq} \left\|W^{L}_{i,:} - W^{L}_{j,:}\right\|_{1}\left\|\rho(\textbf{W}^{L-1}\hat{\textbf{z}}_{\hat{\Wset}}^{L-2}) - \rho(\textbf{W}^{L-1}\textbf{z}_{\hat{\Wset}}^{L-2})\right\|_{\infty} + \epsilon_{N} \left\| W^{L}_{i,:} - W^{L}_{j,:} \right\|_{1} \left\|\textbf{z}_{\Wset}^{N-1}\right\|_{1} \Pi_{k=1}^{L-N-1}\left\|(\textbf{W}^{L-k})^{T}\right\|_{1,\infty} \\
    &\overset{(b)}{\leq} \left\|W^{L}_{i,:} - W^{L}_{j,:}\right\|_{1}\left\|\textbf{W}^{L-1}(\hat{\textbf{z}}_{\hat{\Wset}}^{L-2} - \textbf{z}_{\hat{\Wset}}^{L-2})\right\|_{\infty}  + \epsilon_{N} \left\| W^{L}_{i,:} - W^{L}_{j,:} \right\|_{1} \left\|\textbf{z}_{\Wset}^{N-1}\right\|_{1} \Pi_{k=1}^{L-N-1}\left\|(\textbf{W}^{L-k})^{T}\right\|_{1,\infty}\\
    &\overset{(c)}{\leq}\left\|W^{L}_{i,:} - W^{L}_{j,:}\right\|_{1}\left\|(\textbf{W}^{L-1})^{T}\right\|_{1,\infty}\left\|(\hat{\textbf{z}}_{\hat{\Wset}}^{L-2} - \textbf{z}_{\hat{\Wset}}^{L-2}))\right\|_{\infty} + \epsilon_{N} \left\| W^{L}_{i,:} - W^{L}_{j,:} \right\|_{1} \left\|\textbf{z}_{\Wset}^{N-1}\right\|_{1} \Pi_{k=1}^{L-N-1}\left\|(\textbf{W}^{L-k})^{T}\right\|_{1,\infty}\\
    &\overset{(d)}{\leq} \left\|W^{L}_{i,:} - W^{L}_{j,:}\right\|_{1}\left\|(\textbf{W}^{L-1})^{T}\right\|_{1,\infty}...\left\|(\textbf{W}^{N+1})^{T}\right\|_{1,\infty}\left\| \hat{\textbf{W}}^{N}(\hat{\textbf{z}}_{\hat{\Wset}}^{N-1} - \textbf{z}_{\hat{\Wset}}^{N-1})\right\|_{\infty}  + \epsilon_{N} \left\| W^{L}_{i,:} - W^{L}_{j,:} \right\|_{1} \left\|\textbf{z}_{\Wset}^{N-1}\right\|_{1} \Pi_{k=1}^{L-N-1}\left\|(\textbf{W}^{L-k})^{T}\right\|_{1,\infty}\\
    &\overset{(e)}{\leq} \norm{W^{L}_{i,:} - W^{L}_{j,:}}{1}\Pi_{k=1}^{L-N-1}\norm{\W^{L-k}}{\infty}\big\{ \eps{N}\norm{\z{N-1}{\Wset}}{1} + \eps{\x} \Pi_{m = 1}^{N-1} \norm{\W^{m}}{\infty}(\norm{\W^{N}}{\infty}+ d_{N}\eps{N})\big\}  ,
\end{align}
where inequality (a) results from applying Hölder inequality, and inequality (b) comes from the contractive property (1-Lipschitz) of activation function $\rho(\cdot)$. Inequality (c) and (d) come from triangle inequality applied element-wise on vector $\textbf{W}^{L-1}(\hat{\textbf{z}}_{\hat{\Wset}}^{L-2} - \textbf{z}_{\hat{\Wset}}^{L-2})$ combined with iteration while inequality (e) comes from the constraint of $\eps{N}$ and $\eps{\x}$

With analogous analysis, we proof the event when $N = L$ as following
\begin{align}
     &f^{ij}_{\widehat{\boldsymbol{W}}}(\textbf{x}) - f^{ij}_{\boldsymbol{W}}(\textbf{x}) \nonumber \\
    &\overset{(i)}{\leq} \norm{W^{L}_{i,:} - W^{L}_{j,:}}{1}\norm{\rho(\W^{L-1}\zhat{L-2}{\hat{\Wset}})- \rho(\W^{L-1}\z{L-2}{\Wset})}{\infty} + 2\eps{L}\norm{\rho(\W^{L-1}\zhat{L-2}{\hat{\Wset}})}{1}\\ 
    &\overset{(ii)}{\leq} \eps{\x}\norm{W^{L}_{i,:} - W^{L}_{j,:}}{1}\Pi_{m=1}^{L-1}\norm{\W^{m}}{\infty} +  2\eps{L}\Pi_{m=1}^{L-1}\norm{\W^{m}}{1}(\norm{\x}{1} + d_{0}\eps{\x})
\end{align}
where inequality $(i)$ comes from problem definition (within element-wise $\ell_{\infty}$ norm ball) and since the activation function $\rho(\cdot)$ is non-negative, we could transform the inner product to its $\ell_{1}$ norm. Additionally, inequality $(ii)$ can be easily derived from iterating through the weight matrices and applying the setting of input perturbation $\eps{\x}$

\subsection{Theorem 2: Multi-Layer Scenario} \label{appx:a.2}
In the following proof for Theorem \ref{thm:multi_layer}, we apply similar steps in Appendix \ref{appx_a.1}, introduce one lemma in order to help with the proof of Theorem 2 and consider the difference between set of pairwise margin under natural and weight perturbation setting. Firstly, we have the following lemma for weight perturbation. 

\begin{lemma}[Perturbation of Pure Weight]\label{lem_multiweight}
\label{thm_all_perturb}
Let $f_{\boldsymbol{W}}(\textbf{x}) = \textbf{W}^{L}(...\rho(\textbf{W}^{1}\textbf{x})...)$  denote an $L$-layer (natural) neural network and let   $f_{\boldsymbol{\widehat{W}}}(\textbf{x}) = \hat{\textbf{W}}^{L}(..\hat{\textbf{W}}^{N}...\rho(\hat{\textbf{W}}^{1}\textbf{x})...)$ with $\hat{\textbf{W}}^{k} \in {\rm I\!B}_{\textbf{W}^{k}}^{\infty}(\epsilon_{k}),\ \forall k \in [L]$, denote its perturbed version.
For any set of pairwise margin  $f^{ij}_{\boldsymbol{\widehat{W}}}(\textbf{x})$ and $f^{ij}_{\boldsymbol{W}}(\textbf{x})$, we have
\begin{align*}
 f^{ij}_{\boldsymbol{\widehat{W}}}(\textbf{x})  &\leq f^{ij}_{\boldsymbol{W}}(\textbf{x}) + \left\|W^{L}_{i,:} - W^{L}_{j,:} \right\|_{1} \bigg\{ \epsilon_{1}\left\| \textbf{x} \right\|_{1} \Pi_{l = 1}^{L-2} \left\|(\textbf{W}^{L-l})\right \|_{\infty} \nonumber \\
    &+  \sum_{k = 1}^{L-3} \big (\Pi_{m = k+2}^{L-1} \left\| \textbf{W}^{m} \right \|_{\infty} \big)\epsilon_{k+1} \left\| \textbf{h}^{k^*} \right\|_{1}
    + \epsilon_{L-1} \left\| \textbf{h}^{{L-2}^*} \right\|_{1} \bigg\} + 2\epsilon_{L}\left\| \textbf{h}^{{L-1}^*} \right\|_{1} \nonumber \\
    &= f^{ij}_{\Wset}(\x) + \zeta^{ij}_{\Wset}(\x, \xi) \\
    &\text{where} \hspace{5pt} \textbf{h}^{k^*} = \rho(\textbf{W}^{k^*}...\rho(\textbf{W}^{1^*}
    \textbf{x}) \nonumber \\  
    &\text{with}
    \begin{cases}
        W^{m^*}_{i,j} = W^{m}_{i,j} + \epsilon_{m},\ \forall i, j \ \text{and}~\forall m \in [L] \setminus \{1\} \vspace{5pt} \\
        W^{1^*}_{i,j} = W^{1}_{i,j} + sgn([\textbf{x}]_{j})\hspace{2pt}\epsilon_1, \ \forall i, j
    \end{cases}
\end{align*}%
\end{lemma}
\textit{Proof}: \\ 
\begin{align}
    &f^{ij}_{\widehat{\boldsymbol{W}}}(\textbf{x}) - f^{ij}_{\boldsymbol{W}}(\textbf{x}) \nonumber \\
    &= \{ \hat{W}^{L}_{i,:} - \hat{W}^{L}_{j,:}\}\hat{\textbf{h}}^{L-1} - \{ W^{L}_{i,:} - W^{L}_{j,:}\}\textbf{h}^{L-1} \\
    &\overset{(a)}{\leq} \left\|W^{L}_{i,:} - W^{L}_{j,:}\right\|_{1}\left\|\rho(\hat{\textbf{W}}^{L-1}\hspace{2pt}\hat{\textbf{h}}^{L-2}) - \rho(\textbf{W}^{L-1}\textbf{h}^{L-2})\right\|_{\infty} + 2\epsilon_{L} \boldsymbol{1}^{T}\hat{\textbf{h}}^{L-1}\\
    &\overset{(b)}{\leq} \left\|W^{L}_{i,:} - W^{L}_{j,:}\right\|_{1} \big\{\left\| \textbf{W}^{L-1}(\hat{\textbf{h}}^{L-2} - \textbf{h}^{L-2})\right\|_{\infty} + \left\|(\hat{\textbf{W}}^{L-1} - \textbf{W}^{L-1})\hat{\textbf{h}}^{L-2}\right\|_{\infty} \big\} + 2\epsilon_{L}\left\| \hat{\textbf{h}}^{L-1} \right\|_{1}\\
    &\overset{(c)}{\leq} \left\|W^{L}_{i,:} - W^{L}_{j,:}\right\|_{1} \big\{\left\|(\textbf{W}^{L-1})^{T}\right\|_{1,\infty} \left\|\rho(\hat{\textbf{W}}^{L-2}\hspace{2pt}\hat{\textbf{h}}^{L-3}) - \rho(\textbf{W}^{L-2}\textbf{h}^{L-3})\right\|_{\infty} + \epsilon_{L-1}\left\|\hat{\textbf{h}}^{L-2}\right\|_{1} \big\} \nonumber \\
    &~~~+ 2\epsilon_{L}\left\| \hat{\textbf{h}}^{L-1} \right\|_{1} \\
    &\overset{(d)}{\leq}  \left\|W^{L}_{i,:} - W^{L}_{j,:} \right\|_{1} \bigg\{ \epsilon_{1}\left\| \textbf{x} \right\|_{1} \Pi_{l = 1}^{L-2} \left\|(\textbf{W}^{L-l})^{T}\right \|_{1,\infty} + \sum_{j = 1}^{L-3} \big (\Pi_{k = j+2}^{L-1} \left\| (\textbf{W}^{k})^{T} \right \|_{1,\infty} \big)\epsilon_{j+1} \left\| \hat{\textbf{h}}^{j} \right\|_{1} \nonumber \\
    &~~~+ \epsilon_{L-1} \left\| \hat{\textbf{h}}^{L-2} \right\|_{1}\bigg\} + 2\epsilon_{L}\left\| \hat{\textbf{h}}^{L-1} \right\|_{1} \\
    &\overset{(e)}{\leq}  \left\|W^{L}_{i,:} - W^{L}_{j,:} \right\|_{1} \bigg\{ \epsilon_{1}\left\| \textbf{x} \right\|_{1} \Pi_{l = 1}^{L-2} \left\|(\textbf{W}^{L-l})^{T}\right \|_{1,\infty} + \sum_{j = 1}^{L-3} \big (\Pi_{k = j+2}^{L-1} \left\| (\textbf{W}^{k})^{T} \right \|_{1,\infty} \big)\epsilon_{j+1} \left\| \textbf{h}^{j^*} \right\|_{1} \nonumber \\
    &~~~+ \epsilon_{L-1} \left\| \textbf{h}^{{L-2}^*} \right\|_{1}\bigg\} + 2\epsilon_{L}\left\| \textbf{h}^{{L-1}^*} \right\|_{1}
\end{align} 

In the above proof for lemma, inequality (a) comes from the problem definition and (b) stems from the contractive property of $\rho(\cdot)$ combined with triangle inequality. One could achieve (c) through triangle inequality. By induction and maximization of the $\ell_{1}$ norm of perturbed output under weight perturbation $\hat{\textbf{z}}^{k}$, we could attain inequality (d) and (e).

Thus for any set of pairwise margin  $f^{ij}_{\boldsymbol{\widehat{W}}}(\hat{\textbf{x}})$ and $f^{ij}_{\boldsymbol{W}}(\textbf{x})$, we have
\begin{align}
    &f^{ij}_{\widehat{\boldsymbol{W}}}(\xhat) - f^{ij}_{\boldsymbol{W}}(\textbf{x}) \nonumber \\
    &\overset{(a)}{\leq} \left\|\hat{W}^{L}_{i,:} - \hat{W}^{L}_{j,:}\right\|_{1}\left\|\rho(\hat{\textbf{W}}^{L-1}\hspace{2pt}\zhat{L-2}{\hat{\Wset}}) - \rho(\hat{\textbf{W}}^{L-1}\z{L-2}{\hat{\Wset}})\right\|_{\infty} +\zeta^{ij}_{\Wset}(\x, \xi) \\
    &\overset{(b)}{\leq} \norm{\hat{W}^{L}_{i,:} - \hat{W}^{L}_{j,:}}{1} \big\{ \Pi_{m=1}^{L-1} \norm{\What^{m}}{\infty} \eps{\x}\big\} + \zeta^{ij}_{\Wset}(\x, \xi) \\ 
    &\overset{(c)}{\leq} \eps{\x}\bigg(\norm{W^{L}_{i,:} - W^{L}_{j,:}}{1} + 2d_{L}\eps{L}\bigg)\Pi_{m = 1}^{L-1}(\norm{\W^{m}}{\infty} + d_{m}\eps{m}) +\zeta^{ij}_{\Wset}(\x, \xi) \\ 
    &:= \tau^{ij}_{\Wset}(\xi) + \zeta^{ij}_{\Wset}(\x, \xi)
\end{align} 
Inspecting the above proof, inequality (a) results from separating and applying Lemma \ref{lem_multiweight} and Hölder Inequality. In the other hand, by iterating through the perturbed matrix one could derive inequality (b). Lastly, by applying the constraints on perturbation radius $\eps{m}$ for all layer $m$ and $\eps{\x}$ for the input, we would arrive at the results. 
\end{document}